\documentclass{article}

\usepackage{arxiv}

\usepackage[utf8]{inputenc} % allow utf-8 input
\usepackage[T1]{fontenc}    % use 8-bit T1 fonts
\usepackage{url}            % simple URL typesetting
\usepackage{booktabs}       % professional-quality tables
\usepackage{amsfonts}       % blackboard math symbols
\usepackage{nicefrac}       % compact symbols for 1/2, etc.
\usepackage{microtype}      % microtypography
\usepackage{lipsum}

\usepackage{thmtools, thm-restate}
\usepackage{hyperref}
\usepackage{times}
\usepackage{latexsym}
\usepackage{amsmath}
\DeclareMathOperator*{\argmax}{argmax} % thin space, limits
\usepackage{bbm}
\usepackage{multirow}
\usepackage{graphicx}
\usepackage{algorithm}
\usepackage[noend]{algpseudocode}
\usepackage{amsfonts}
\usepackage{tabularx}
\usepackage{booktabs,caption}
\usepackage[flushleft]{threeparttable}

\declaretheorem{theorem}
\declaretheorem{proposition}

\makeatletter
\def\BState{\State\hskip-\ALG@thistlm}
\makeatother

\makeatletter
\newcommand{\fixed@sra}{$\vrule height 2\fontdimen22\textfont2 width 0pt\shortrightarrow$}
\newcommand{\shortarrow}[1]{%
  \mathrel{\text{\rotatebox[origin=c]{\numexpr#1*45}{\fixed@sra}}}
}
\makeatother

\title{A Label Proportions Estimation Technique for Adversarial Domain Adaptation in Text Classification}

\author{
 Zhuohao Chen \\
  Signal Analysis and Interpretation Lab\\
  University of Southern California
   \And
 Karan Singla \\
  Signal Analysis and Interpretation Lab\\
  University of Southern California\\
  \And
 David C. Atkins \\
  Department of Psychiatry and Behavioral Sciences\\
  University of Washington\\
  \And
 Zac E Imel \\
  Department Educational Psychology\\
  University of Utah\\
  \And
 Shrikanth Narayanan \\
  Signal Analysis and Interpretation Lab\\
  University of Southern California\\
}

\begin{document}
\maketitle
\begin{abstract}
Many text classification tasks are domain-dependent, and various domain adaptation approaches have been proposed to predict unlabeled data in a new domain. Domain-adversarial neural networks (DANN) and their variants have been used widely recently and have achieved promising results for this problem. However, most of these approaches assume that the label proportions of the source and target domains are similar, which rarely holds in most real-world scenarios. Sometimes the label shift can be large and the DANN fails to learn domain-invariant features. In this study, we focus on unsupervised domain adaptation of text classification with label shift and introduce a domain adversarial network with label proportions estimation (DAN-LPE) framework. The DAN-LPE simultaneously trains a domain adversarial net and processes label proportions estimation by the confusion of the source domain and the predictions of the target domain. Experiments show the DAN-LPE achieves a good estimate of the target label distributions and reduces the label shift to improve the classification performance. 
\end{abstract}

% keywords can be removed

\section{Introduction}
Text classification is one of the most important tasks in natural language processing (NLP). However, many text data sets are unlabelled. Moreover, text data is always domain-dependent and it is difficult to obtain annotated data for all the domains of interest. To handle this, researchers use domain adaptation techniques for text classification. For example,  \cite{blitzer2007biographies} first applied  structural correspondence learning (SCL) \cite{blitzer2006domain} to cross-domain sentiment classification. In \cite{pan2010cross} spectral feature alignment (SFA) was proposed to reduce the gap between domain-specific words of the domains. The study \cite{bollegala2015cross} modeled the cross-domain classification task as an embedding learning. 

Recently deep adversarial networks \cite{goodfellow2014generative} have achieved success across many tasks including text classification. The domain-adversarial neural networks (DANN) structure proposed by \cite{ganin2016domain} outperforms the traditional approaches in domain adaptation tasks of sentiment analysis. It implements a domain classifier to learn domain-invariant features. The study \cite{chen2018adversarial} applied the adversarial deep networks to the cross-lingual sentiment classification. Other studies extended DANN for different multi-source scenarios \cite{liu2017adversarial, zhao2018multiple,chen2018multinomial}. However, they all assume that the label proportions across the domains remain unchanged, an assumption that often is not met in real world tasks. 

The changes in the label distribution are known as prior probability shift or label shift which prohibit the DANN from learning domain-invariant features\cite{zhao2019learning}. To estimate the label shift, \cite{saerens2002adjusting} proposed an EM algorithm to obtain the new a priori probabilities by maximizing the likelihood of the new data. This approach has been successfully applied in \cite{chan2006estimating}. In the study of \cite{zhang2013domain} the kernel mean matching (KMM) method was demonstrated to correct the shift. \cite{nguyen2016continuous} further developed the KMM algorithm for continuous target shift adaptation. A recent attempt to quantify the shift is the Black Box Shift Estimation (BBSE) \cite{lipton2018detecting, azizzadenesheli2019regularized}, a moment-matching approach using confusion matrices which achieves accurate estimates on high-dimensional datasets. However, these approaches are under an anticausal hypothesis in which the labels cause the features \cite{scholkopf2012causal}. %Recently, \cite{li2019target} proposed to address the label shift problem by using distribution matching to estimate label proportions. 

In this paper, we implement a domain adversarial network framework with label proportions estimation (DAN-LPE) which learns domain-invariant features and estimates the target label proportions. The proportion estimation only uses the confusion and target label predictions as inputs. To reduce the label shift , we apply the similar trick in  \cite{li2019target} by re-weighting the sample in the domain classifier based on the estimated label distribution. In the experiments on two data sets (Yelp dataset, behavioral coding of psychotherapy conversations), we compare DAN-LPE with other algorithms in terms of the label distribution estimates and classification performance and show that it leads in most of the tasks.

\section{Problem Setup}

Let $P$ and $Q$ be the source and target domains defined on $X\times Y$ and $f$ be any text classifier. We use $x \in X = R^d$ and $y \in Y = \{1,2,...,L\}$ to denote the feature and label variables.  The output of the classifier is denoted by $\hat{y}=f(x)$. We use $p$ and $q$ to indicate the probability density functions of $P$ and $Q$, respectively. The source and target datasets are represented by $D^P = \{(x^P_1, y^P_1), (x^P_2, y^P_2),...,(x^P_M, y^P_M)\}$ and $D^Q = \{(x^Q_1, y^Q_1), (x^Q_2, y^Q_2),...,(x^Q_N, y^Q_N)\}$. We split $D^P$ into the training set $D^{P_t}$ and validation set $D^{P_v}$. The prior distributions of $P$ and $Q$ are given by $\alpha_l = p(y=l)$ and $\beta_l = q(y=l)$.

\begin{figure}[htb]
  \centering
  \includegraphics[width=10cm]{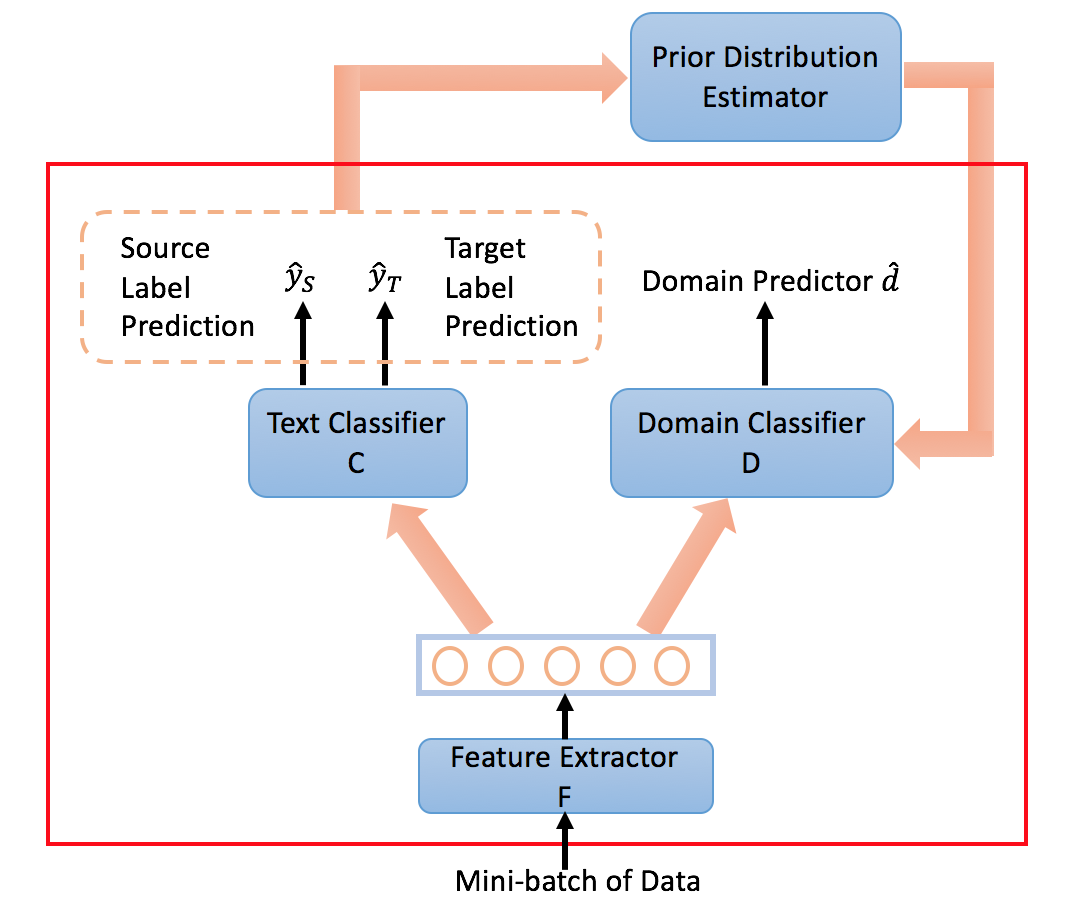}
  \caption{The Structure of DAN-LPE}
  \label{fig:dan_lpe}
\end{figure}

%$X^p = {(x^p_1, y^p_1), (x^p_1, y^p_1),...,(x^p_M, y^p_M)}$ 
The red box in Fig.~\ref{fig:dan_lpe} shows the DANN structure consisting of a feature extractor $F$, a text classifier $C$ and a domain classifier $D$. We expect the feature extractor to capture the features satisfying $p(x|y) = q(x|y)$ with the help of $D$ which makes the feature distributions between source and target domains indistinguishable by back-propagation with gradient reversal. However, the performance of $D$ would be declined if the prior distributions between $P$ and $Q$ differ a lot. To handle this, we implement a prior distribution estimator above the red box in Fig.~\ref{fig:dan_lpe} to estimate the target label proportions and correct the label shift by re-weighting the samples feeding into the $D$ based on the varying estimated proportions. The prior distribution estimator takes the confusion of $D^{P_t}$ and the label predictions of $D^Q$ as inputs. 

\section{Domain Adversarial Network With Label Proportion Estimation}
\subsection{Moments and Matrices Definition}
We first define the training set $D^{P_T} = \{(x^{P_T}_1, y^{P_T}_1), (x^{P_T}_2, y^{P_T}_2),...,(x^{P_T}_{M_T}, y^{P_T}_{M_T})\}$.
The moments, matrices of $p$ and $q$ and denoted as follows

\begin{gather*} % the "starred" equation environments produce no equation numbers
P_{ij}=p(y=i,\hat{y}=j)\;\;\;\; Q_{ij}=q(y=i,\hat{y}=j)\\
%\hat{P}_{ij}=\frac{1}{M_T}\sum_{k=1}^{M_T}\mathbbm{1}\{y=i,f(x_k^{P_T})=j\}\\
%\hat{Q}_{ij}=\frac{1}{N}\sum_{k=1}^{N}\mathbbm{1}\{y=i,f(x_k^{Q})=j\}\\
P_i^j=p(\hat{y}=j \mid y=i) = \frac{p(y=i, \hat{y}=j)}{p(y=i)} = \frac{P_{ij}}{\sum_k{P_{ik}}}\\
Q_i^j=q(\hat{y}=j \mid y=i) = \frac{q(y=i, \hat{y}=j)}{q(y=i)} = \frac{Q_{ij}}{\sum_k{Q_{ik}}}\\
%\hat{P}_i^j=\frac{\hat{P}_{ij}}{\sum_k{\hat{P}_{ik}}} \;\;\;\;\;\; \hat{Q}_i^j=\frac{\hat{Q}_{ij}}{\sum_k{\hat{Q}_{ik}}}
\end{gather*}

We also present the plug-in estimates using the samples from $D^{P_T}$ and $D^Q$

\begin{gather*} % the "starred" equation environments produce no equation numbers
\hat{P}_{ij}=\frac{1}{M_T}\sum_{k=1}^{M_T}\mathbbm{1}\{y_k^{P_T}=i,f(x_k^{P_T})=j\}\\
\hat{Q}_{ij}=\frac{1}{N}\sum_{k=1}^{N}\mathbbm{1}\{y_k^Q=i,f(x_k^{Q})=j\}\\
\hat{P}_i^j= \frac{\sum_{k=1}^{M_T}\mathbbm{1}\{y_k^{P_T}=i,f(x_k^{P_T})=j\}}{\sum_{k=1}^{M_T}\mathbbm{1}\{y_k^{P_T}=i\}}=\frac{\hat{P}_{ij}}{\sum_k{\hat{P}_{ik}}}\\
\hat{Q}_i^j= \frac{\sum_{k=1}^{N}\mathbbm{1}\{y_k^{Q}=i,f(x_k^{Q})=j\}}{\sum_{k=1}^{N}\mathbbm{1}\{y_k^{Q}=i\}}=\frac{\hat{Q}_{ij}}{\sum_k{\hat{Q}_{ik}}}
\end{gather*}

\begin{proposition}
\label{thm:thm2}
Assume $p(y=i) > 0$,  $q(y=i) > 0$ holds $\forall i$, then $\hat{P}_{ij} \xrightarrow{a.s.} P_{ij}$, $\hat{Q}_{ij} \xrightarrow{a.s.} Q_{ij}$, $\hat{P}_i^j  \xrightarrow{a.s.} P_i^j$ and $\hat{Q}_i^j  \xrightarrow{a.s.} Q_i^j $ as $M_T$,  $N \to \infty$.
\end{proposition}

The proof of Proposition~\ref{thm:thm2} is demonstrated in the Appendix using the Law of Large Numbers(LLN).\\

We cannot obtain $\hat{Q}_{ij}$ and $\hat{Q}_i^j$ because $Q$ is an unlabeled domain. However, the distribution of the label predictions of $Q$ is accessible.

\begin{gather*}
\beta_i = q(y=i) =\sum_{j=1}^{L}q(y=i,\hat{y}=k)= \sum_{j=1}^{L}{Q_{ij}}\\
q(\hat{y}=j)=\sum_{i=1}^{L}q(y=i,\hat{y}=j)=\sum_{i=1}^{L}Q_{ij}=\sum_{i=1}^{L}\beta_{i}Q_i^j
\end{gather*}

We can get the estimate of $q(\hat{y}=j)$ by

\begin{gather*}
\hat{q}(\hat{y}=j)=\frac{1}{N}\sum_{k=1}^{N}\mathbbm{1}\{f(x_k^{Q})=j\} = \frac{1}{N}\sum_{k=1}^{N}\sum_{i=1}^{L}\mathbbm{1}\{y_k^Q=i,f(x_k^{Q})=j\}=\sum_{i=1}^{L}\hat{Q}_{ij}
\end{gather*}

By Proposition~\ref{thm:thm2} we can also conclude that $\hat{q}(\hat{y}=j) \xrightarrow{a.s.} q(\hat{y}=j)$.

\subsection{Label Proportions Estimation}

The crucial component of the DAN-LPE is the way of updating label proportion estimates. We define a random vector $\gamma = [\gamma_1, \gamma_2,...,\gamma_L]^T$ to be the estimator of $\beta$. Once the perfect domain-invariant features are learnt such that $p(x|y) = q(x|y)$, which implies $P_i^j=Q_i^j$, and the target label proportions are accurately estimated that  $\gamma=\beta$, in this condition, the equality $\sum_{i=1}^{L}\gamma_{i}P_i^j=\sum_{i=1}^{L}\beta_{i}Q_i^j=q(\hat{y}=j)$ holds for every $j$. So we proposed the following loss function

\begin{equation}
J_{\gamma}=\sum_{j=1}^{L}(\sum_{i=1}^{L}\gamma_{i}P_i^{j}-q(\hat{y}=j))^2\label{eq:1}
\end{equation}

Replacing with the plug-in estimates we get

\begin{equation}
\hat{J_{\gamma}}=\sum_{j=1}^{L}(\sum_{i=1}^{L}\gamma_{i}\hat{P}_i^{j}-\hat{q}(\hat{y}=j))^2\label{eq:2}
\end{equation}

%Now we relate $\gamma$ only to the observable data. 
We set $J_{\gamma}=0$, which implies $\sum_{i}\gamma_{i}P_i^j=q(\hat{y}=j), \forall j$, and we get that

\begin{equation}\label{eq:3}
\begin{split}
\bar{P}\cdot\gamma & = [q(\hat{y}=1),q(\hat{y}=1),...,q(\hat{y}=L)]^T\\
\bar{P}&=
  \begin{bmatrix}
   P_1^1  & P_1^2  & \dots & P_L^1 \\
   P_2^1  & P_2^2  & \dots & P_L^2 \\
   \vdots  & \vdots  & \ddots & \vdots \\
   P_1^L  & P_2^L  & \dots & P_L^L \\ 
   \end{bmatrix}
\end{split}
\end{equation}

Then we conclude the following implication

\begin{theorem}
\label{thm:thm1}
Assume $p(x|y) = q(x|y)$ and $\bar{P}$ is an invertible matrix, then $J_{\gamma} \geq 0$ is a convex function of $\gamma$ and the equality is satisfied when $\gamma=\beta$.
\end{theorem}

By computing the gradient we derive that
\begin{equation}
\frac{\partial J_{\gamma}}{\partial\gamma_k}=2\sum_{j=1}^{L}P_k^j(\sum_{i=1}^{L}\gamma_{i}P_i^{j}-q(\hat{y}=j))\label{eq:4}
\end{equation}

We modify the equation in a similar way as previous and relate $\gamma$ only to the observable data

\begin{equation}
\frac{\partial\hat{J_{\gamma}}}{\partial\gamma_k}=2\sum_{j=1}^{L}\hat{P}_k^j(\sum_{i=1}^{L}\gamma_{i}\hat{P}_i^{j}-\hat{q}(\hat{y}=j))\label{eq:5}
\end{equation}

 The proposed prior $\gamma$ is updated by gradient descent using Equation (\ref{eq:5}). However, since $\gamma$ is constrained by $\sum_{i}\gamma_{i}=1$, we apply the projected gradient descent instead

\begin{equation}
\begin{split}
G(\gamma)&=\frac{\partial\hat{J_{\gamma}}}{\partial\gamma}= [\frac{\partial\hat{J_{\gamma}}}{\partial\gamma_1}, \frac{\partial\hat{J_{\gamma}}}{\partial\gamma_2},...,\frac{\partial\hat{J_{\gamma}}}{\partial\gamma_L}]^T\\
\gamma^{t}&=\gamma^{t-1} - \lambda_L (G(\gamma^{t-1}) - <G(\gamma^{t-1}),\frac{\mathbbm{1}^T}{\sqrt{L}}>\cdot\frac{\mathbbm{1}}{\sqrt{L}})\label{eq:6}
\end{split}
\end{equation}

Where $\lambda_L$ is the learning rate of updating $\gamma$. To avoid the existence of the negative proportion estimate, we also set a lower bound that $\gamma_i \geq 0.001$. Once $\gamma_i < 0.001$, we have

\begin{equation}
\begin{split}
\gamma_k &= \gamma_k + \gamma_i - 0.001, k = \argmax\limits_j \gamma_j \\
\gamma_i &= 0.001 \label{eq:7}
\end{split}
\end{equation}

We define $J_C$ and $J_D$ as the loss functions of $C$ and $D$. To eliminate the prior shift, we re-weight the samples from $P$ in $D$ based on their labels. Let $w_i=\frac{\gamma_i}{\tilde{\alpha}_i}$, where $\tilde{\alpha_i}$ is the prior distribution of $D^{P_T}$. For a mini-batch of size $B$, the instances from $P$ and $Q$ are $B^P=\{(\mu_1,\nu_1),...,(\mu_{B/2},\nu_{B/2})\}$ and $B^Q=\{(\mu'_1,\nu'_1),...,(\mu'_{B/2},\nu'_{B/2})\}$, the sample weight vector of $B^P$ is $w^T = [w_1,w_2,...,w_{B/2}]$. And we compute $J_D$ by

\begin{equation}
 \begin{split}
J_D&=\frac{1}{B/2}(\sum_{i=1}^{B/2}\frac{w_{v_i}}{\Vert w\Vert}c(\mu_i) + \sum_{i=1}^{B/2}c(\mu'_i))\label{eq:8}
\end{split}
\end{equation}

Where $c(\cdot)$ presents the cross-entropy loss. By this new loss function, the samples from the same class in the source and target domain have closer contributions in $D$, which helps the domain adapter to suffer less from the label shift.

\begin{algorithm}[htb]
 \caption{Domain Adversarial Network with Label Proportion Estimation}\label{alg:alg1}
 \begin{algorithmic}[1]
 \State \textbf{Step 1:}
 \State Initialization: $\gamma = [\frac{1}{L},\frac{1}{L},...,\frac{1}{L}]^T$; $\lambda_D > 0$, $\lambda_L > 0$; $T, T_0, k, m \in \mathbb{N}$
   \For{$t$ = $i$ to $T$}
   \State Sample a mini-batch training set for $C$ and $D$ respectively
   \State Fix $\gamma$
   \State Update C parameters using $\nabla J_C$
   \State Update F parameters using $\nabla J_C - \lambda_D \nabla J_D$
   \State Update D parameters using $\nabla J_D$
   \If{$t > T_0$ and $t$ mod $k =0$}
   \State Predict the labels of $D^{P_T}$ and $D^Q$.
   \For{$j$ = 1 to $m$}
   \State Update $\gamma$ by Equation (\ref{eq:5})-(\ref{eq:7})
   \EndFor
   \EndIf
   \EndFor
   \State
   \State \textbf{Step 2:}
   \State Perform DANN with fixed $\gamma$ and modified loss function of $D$ in Equation (\ref{eq:8})
\end{algorithmic}
\end{algorithm}

The complete pseudo-code of this learning procedure is given in Algorithm \ref{alg:alg1}. In the first step it trains a domain adversarial network and processes label proportion estimation alternately to get an estimate of the target prior distribution. During this procedure, we are achieving a more and more accurate estimate of the target label proportions, the label shift effect is being reduced by re-weighting the samples in $D$ and better domain-invariant features are learnt. Since the label shift still matters a lot in early epochs, we need a second step to perform general DANN with the fixed $\gamma$ achieved in the first step and the modified loss function $J_D$ in (\ref{eq:6}).

The hyper-parameters of step 1 in algorithm \ref{alg:alg1} are quite flexible. The number of iterations $T$ is deemed adequate when the validation loss does not increase measurably. The role of $T_0$ is to guarantee that we update $\gamma$ when a decent model is trained. We update $\gamma$ every $k$ iterations so it reduces the times to predict $D^{P_T}$ and $D^{Q}$ and accelerates the process. Parameter $\lambda_L$ and $m$ controls how fast  and smoothly $\gamma$ changes. The DAN-LPE is not very sensitive to these hyper-parameters. When $\gamma$ is fixed as the prior distribution of $D^{P_T}$, step 1 of Algorithm \ref{alg:alg1} is equivalent to the basic DANN.

\section{Experiments}

In this section, we perform the experiments on two different data sets and show the classification results of different models. We also present the label distribution estimates and compare it with the BBSE outcomes. 

\subsection{Experiments on Yelp Data}

The Yelp Open Dataset \cite{yelpdata2019} includes 192,609 businesses and 6,685,900 reviews of more than 20 categories. In each review a user expresses opinions about a business and gives a rating ranging from 1 to 5. We compute the average review ratings $z_i$ of each business and label the business with

\begin{equation}
  y=\begin{cases}
    1, & \text{if $z_i<3.4$}.\\
    0, & \text{if $z_i>3.6$}.
  \end{cases}\label{eq:9}
\end{equation}

%$y=0$ if $z_i<3.4$ and $y=1$ if $z_i>3.6$. 
 
\begin{figure}[htb]
\centering
\begin{minipage}[b]{.5\linewidth}
  \centering
  \centerline{\includegraphics[width=4.8cm]{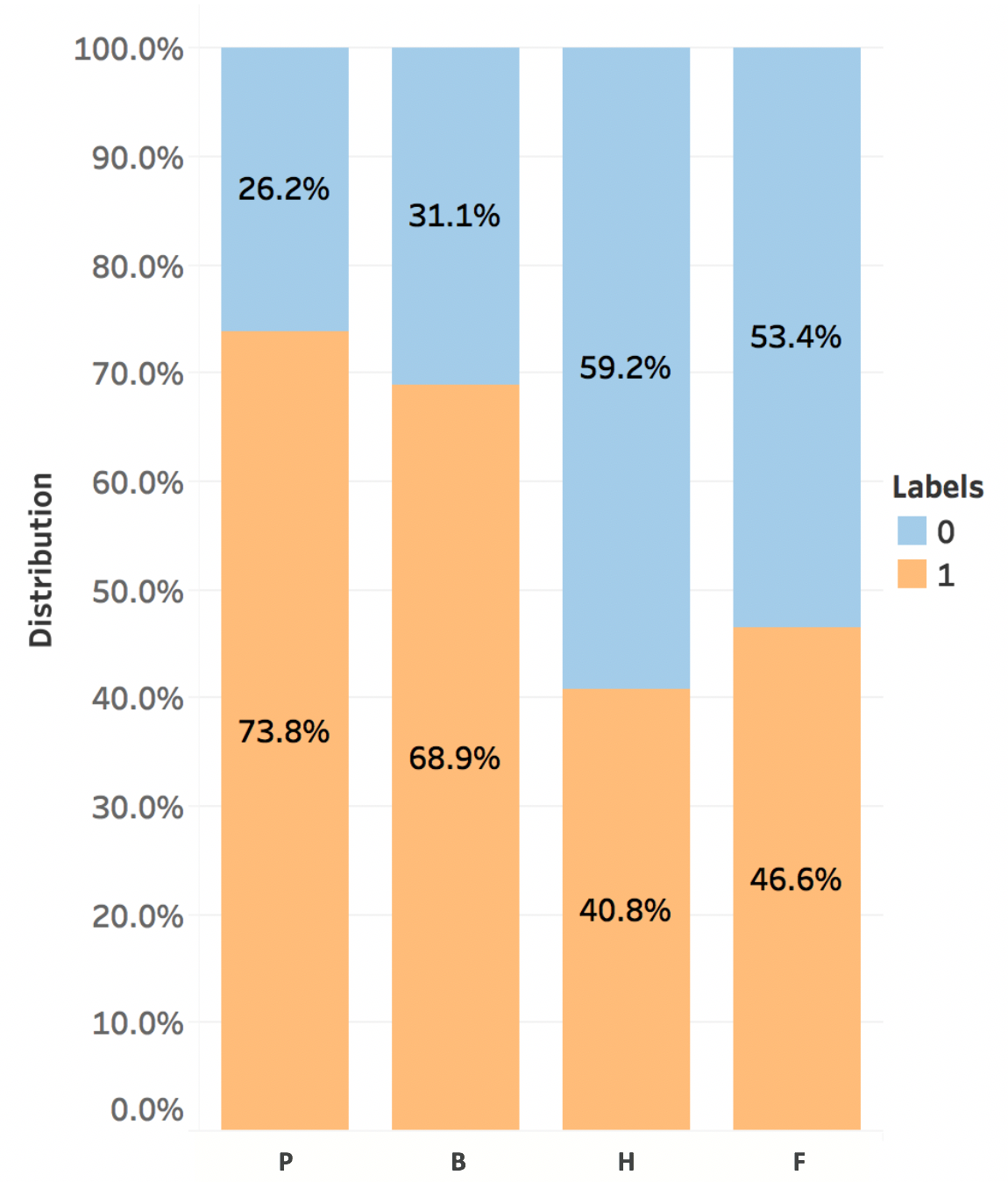}}
%  \vspace{1.5cm}
  \centerline{\footnotesize{(a) Yelp Data}}\medskip
\end{minipage}
\hfill
\begin{minipage}[b]{0.42\linewidth}
  \centering
  \centerline{\includegraphics[width=4.2cm]{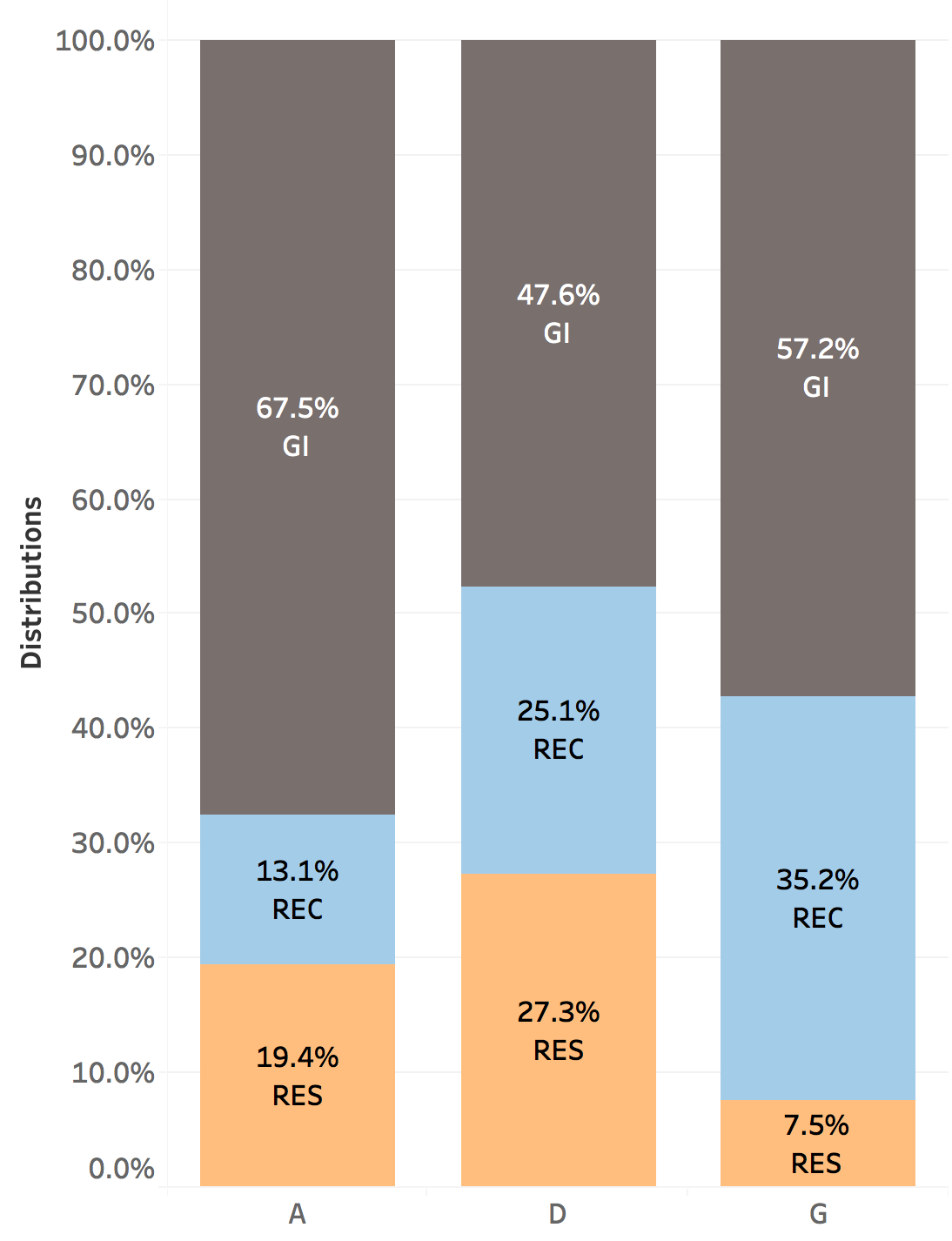}}
%  \vspace{1.5cm}
  \centerline{\footnotesize{(b) Behavioral Coding Data}}\medskip
\end{minipage}
\caption{Label Distributions}
\label{fig:distribution}
\end{figure}

The business with $3.4\leq z_i \leq 3.6$ are filtered out to make the gap. We select the data of Financial Services(F), Hotel$\&$Travel(H), Beauty$\&$Spas(B) and Pets(P) for the tasks. Their label distributions vary as shown in Fig.~\ref{fig:distribution}a. We sample 2800 businesses for each domain preserving the label proportions and predict the class using their reviews. Among the samples of each domain, 10\% of them are split into the validation set.

We extract the features for each business using the following steps:\\
1) remove punctuations and the stop words from Natural Language Toolkit (NLTK)\cite{loper2002nltk};\\
2) apply stemming using Porter algorithm implemented in NLTK;\\
3) negate words between the negation and the following punctuation \cite{das2007yahoo};\\
4) find 500 words by the intersection of exact 837 most common words of each domain and form the bag of words representation for each review by the occurrence of these tokens;\\ 
5) compute the averaging the vectors of its reviews to get the feature vector  of this business.

In the DAN-LPE setting we implement a standard neural network with 2 hidden layers of 32 dimensions. $D$ takes the output of the first layer as the input and another hidden layer of the same size. Dropout of p = 0.6 is set for all the hidden layers. We compare DAN-LPE with SVM, DNN and DANN. DNN is constructed by $F+C$ and DANN by $F+C+D$. For DNN, DANN and DAN-LPE, the learning rate is fixed as $10^{-4}$ and the size of mini-batch is 64. For optimization, the Adam \cite{kingma2014adam} optimizer was applied following an early stopping strategy. In the first step of DAN-LPE, we set $\lambda_D=0.05$, $\lambda_L=0.01$, $T=8000$, $T_0=2000$ and $m=k=5$.

To evaluate the label proportions estimation, we define $\hat{\gamma}_{dl}$ to be the estimate using DAN-LPE, $\hat{\gamma}_{b}$ to be the estimate using BBSE and $\hat{\alpha}$ and $\hat{\beta}$ be the label proportions of samples in the source and target dataset. The results are measured by the Euclidean distance between the estimate and the actually label proportions of the target set.

The results are shown in Table \ref{tab:yelp}. We found the DAN-LPE has a overall more accurate label proportions estimate than BBSE. In the first eight tasks $\hat{\alpha}$ and $\hat{\beta}$ differ a lot and the DANN does not show much improvement over DNN. In some tasks it even degrades the classification performance. In these experiments, the DAN-LPE shows a significant gain because the label proportions estimate $\hat{\gamma}$ reduces the label shift. In the last four tasks the label proportions between the source and target domains are close and DANN gets the best accuracy in three of them. However, the DAN-LPE algorithm performs comparably with DANN in these tasks since is does not degrade in estimating $\beta$. It is worth mentioning that given accurate label shift estimates, we can also improve the classification accuracy by the prior probability adjustment \cite{saerens2002adjusting}, re-weighting the class importance in $C$. Here we focus on the behavior in the domain adapter and will not further discuss this aspect in this paper.

\begin{table}[htb]
\centering
\resizebox{0.7\textwidth}{!}{\begin{tabular}{c|cccc|ccc}
\hline
\textbf{Task} & \multicolumn{4}{c}{\textbf{Accuracy}} & \multicolumn{3}{c}{\textbf{Estimation Results}} \\
                   P-\textgreater{}Q   & SVM  & DNN  & DANN & \begin{tabular}[c]{@{}c@{}}DAN-\\ LPE\end{tabular} & $|\hat{\beta} -\hat{\gamma}_{dl}|$            &   $|\hat{\beta} -\hat{\gamma}_{b}|$   & $|\hat{\beta} - \hat{\alpha}|$        \\
\hline
B-\textgreater{}H     & 0.881    & 0.882    & 0.884    & \textbf{0.886}       & \textbf{0.08}     &  0.15           & 0.40            \\
B-\textgreater{}F     & 0.869    & 0.876    & 0.883    & \textbf{0.884}       & \textbf{0.10}     & 0.13            & 0.32
        \\
P-\textgreater{}H     & 0.842    & 0.863    & 0.858    & \textbf{0.865}       & \textbf{0.03}     & 0.06            & 0.47 
        \\
P-\textgreater{}F     & 0.871    & 0.879    & 0.880    & \textbf{0.883}       & \textbf{0.13}     & 0.17            & 0.38
        \\
H-\textgreater{}B     & 0.862    & 0.861    & 0.858    & \textbf{0.868}       & \textbf{0.05}    & \textbf{0.05}    & 0.40
        \\
H-\textgreater{}P     & 0.871    & 0.878    & 0.875    & \textbf{0.879}       & \textbf{0.06}   & 0.08          & 0.47
        \\
F-\textgreater{}B     & 0.885    & 0.879    & 0.877    & \textbf{0.896}       & \textbf{0.03}     & 0.05        & 0.32
        \\
F-\textgreater{}P     & 0.840    & 0.828    & 0.826    & \textbf{0.845}       & \textbf{0.03}       & 0.07      & 0.38
        \\
B-\textgreater{}P     & 0.884    & 0.892    & \textbf{0.893}    & \textbf{0.893}       & \textbf{0.02}      & 0.05       & 0.07
        \\
P-\textgreater{}B     & 0.896    & 0.907    & \textbf{0.908}    & \textbf{0.908}       & \textbf{0.06}      & 0.07       & 0.07
        \\
H-\textgreater{}F     & 0.881    & \textbf{0.885}    & 0.883    & \textbf{0.885}       & \textbf{0.02}    & 0.03        & 0.08
        \\
F-\textgreater{}H     & 0.846    & 0.839    & \textbf{0.852}    & 0.849       & 0.13      &   0.17    & \textbf{0.08}
\\
\hline
\end{tabular}}
\caption{\label{tab:yelp} Results of Yelp Experiments.}
\end{table}

\subsection{Experiments on Behavioral Coding Data from Psychotherapy Conversations}

The other text classification application considered is automated behavioral coding in the context of psychotherapy; specifically we consider data from Motivational Interviewing (MI) \cite{miller2012motivational} counseling sessions. The utterances of therapist therein are coded to evaluate a therapist based on the Motivational Interviewing Skill Code (MISC) \cite{miller2003manual} manual. Some of these MISC codes inherently overlap with each other in their construction, and training classifiers for these confusable codes can help improve the behavioral coding performance \cite{chen2019improving}. In this experiment we classify the utterances of Giving Information (GI), simple reflection (RES) and complex reflection(REC) collected from MI sessions of alcohol addiction (A), drug-abuse (D) \cite{atkins2014scaling} and general psychology conversations (G) from a US university counseling center with each category containing around 10000 samples. The label proportions are shown in Fig.~\ref{fig:distribution}b. 

The modules $C$ and $D$ in the DAN-LPE structure are similar to the one in the Yelp experiment with dimension of 128 in hidden layers and dropout rate p = 0.4. We replace $F$ of the yelp experiment with a word embedding layer, followed by a bidirectional LSTM layer and an attention mechanism implemented as in \cite{yang2016hierarchical} above the LSTM. As shown in Fig.~\ref{fig:distribution}b, the data are highly imbalanced so we evaluate the performance by the average F1 score. In module $C$ of both step 1 and step 2 of the algorithm \ref{alg:alg1}, we assign weights for each class inversely proportional to their class frequencies to make the algorithm more robust as well as for improving the F-score.

The results in Table \ref{tab:BC} show that the DAN-LPE wins BBSE and reduce the label shift in all the tasks and achieves the overall best classification performance. It gains the highest F-score in most tasks except in the last one when the estimated proportions do not decrease the label shift much. The DANN only has a comparable F-score compared with DNN and even degrades for some tasks. The DNN results suggest that we find the behavioral coding task is harder than the yelp experiment because the number of classes is larger and the behavior codes are human defined and not uncorrelated. However, DAN-LPE shows its robustness and still gives reasonable proportion estimate of the data in unlabelled domain.  

\begin{table}[htb]
\centering
\resizebox{0.7\textwidth}{!}{\begin{tabular}{c|ccc|ccc}
\hline
\textbf{Task} & \multicolumn{3}{c}{\textbf{F-score}} & \multicolumn{3}{c}{\textbf{Estimation Results}} \\
                  P-\textgreater{}Q    & DNN    & DANN    & DAN-LPE   & $|\hat{\beta} - \hat{\gamma}_{dl}|$           &  $|\hat{\beta} - \hat{\gamma}_{b}|$   & $|\hat{\beta} - \hat{\alpha}|$              \\
\hline
A-\textgreater{}G     & 0.496      & 0492       & \textbf{0.503}         & \textbf{0.14}    & 0.15        & 0.27              \\
G-\textgreater{}A     & 0.489      & 0.489       & \textbf{0.496}         & \textbf{0.10}      & 0.17        & 0.27             \\
D-\textgreater{}G     & 0.512      & 0.508       & \textbf{0.522}         & \textbf{0.05}       & 0.08         & 0.24   \\
G-\textgreater{}D     & 0.552      & 0.556      & \textbf{0.558}        & \textbf{0.06}     & 0.15           & 0.24   \\
A-\textgreater{}D     & 0.627      & 0.639      & \textbf{0.644}        & \textbf{0.13}     & 0.15           & 0.25   \\
D-\textgreater{}A     & 0.593      & \textbf{0.594}       & 0.593         & \textbf{0.19}       & 0.25         & 0.25    \\
\hline
\end{tabular}}
\caption{\label{tab:BC} Results of Behavioral Coding Experiments.}
\end{table}

\section{Conclusion and Future Work}

In this paper, we proposed the DAN-LPE framework to handle the label shift in DANN for unsupervised domain adaptation of text classification. In DAN-LPE we estimate the target label distribution and learn the domain-invariant features simultaneously. We derived the formula to update the label proportions estimate using the confusion and target label predictions and re-weighted the samples in the domain classifier to learn better domain-invariant features. Experiments shows that the DAN-LPE gives much better estimate than the BBSE and evidently reduces the label shift. When the DANN does not gain much from the domain adapter under large shift, the DAN-LPE structure successfully corrects the shift and achieves better performance. In the future, we plan to apply the DAN-LPE to other tasks such as image classification.

\section*{Acknowledgments}

\bibliography{label-shift-dann}
\bibliographystyle{unsrt}  


\appendix
\section{Proof of Proposition~\ref{thm:thm2} (See page~\pageref{thm:thm2})}

For any $i, j \in {1,2,...,L}$, $\hat{P}_{ij}$ can be considered as mean of independent and identically distributed random variables

\begin{equation} \label{eq:10}
\begin{split}
\hat{P}_{ij}&=\frac{1}{M_T}\sum_{k=1}^{M_T}\mathbbm{1}\{y_k^{P_T}=i,f(x_k^{P_T})=j\} = \frac{1}{M_T}\sum_{k=1}^{M_T}z_k
\end{split}
\end{equation}

Where $\{z_k\}_{k=1}^{M_T}$ are i.i.d random variables bounded by [0, 1] with $\mathbb{E}(z_k) = P_{ij}$ and variance  $ \sigma_z^2< \infty$, thus

By the strong law of large numbers (SLLN), as $M_T \to \infty$, $\frac{1}{M_T}\sum_{k=1}^{M_T}z_k = \hat{P}_{ij} \xrightarrow{a.s.} P_{ij}$. The proof for $\hat{Q}_{ij}$ is similar as $\hat{P}_{ij}$.\\

The expression of $\hat{P}_i^j$ can be replaced by

\begin{equation} \label{eq:11}
\begin{split}
\hat{P}_i^j &= \frac{\frac{1}{M_T}\sum_{k=1}^{M_T}\mathbbm{1}\{y_k^{P_T}=i,f(x_k^{P_T})=j\}}{\frac{1}{M_T}\sum_{k=1}^{M_T}\mathbbm{1}\{y_k^{P_T}=i\}} \\
&= \frac{ \frac{1}{M_T}\sum_{k=1}^{M_T}z_k}{ \frac{1}{M_T}\sum_{k=1}^{M_T}s_k} = \frac{Z_{M_T}}{S_{M_T}}
\end{split}
\end{equation}

Where $\{s_k\}_{k=1}^{M_T}$ are i.i.d random variables bounded by [0, 1] with $\mathbb{E}(s_k) = p(y=i) = \alpha_i > 0$ and variance  $ \sigma_s^2< \infty$, by the SLLN we conclude as $M_T \to \infty$, $Z_{M_T} \xrightarrow{a.s.} P_{ij}$ and $S_{M_T} \xrightarrow{a.s.} \alpha_i$. Then we derive

\begin{equation} \label{eq:12}
\begin{split}
& \mathbbm{P}(\lim_{M_T\to\infty} \frac{Z_{M_T}}{S_{M_T}} \neq \frac{P_{ij}}{\alpha_i}) \leq \mathbbm{P}(\lim_{M_T\to\infty} \{Z_{M_T} \neq P_{ij}\}\cup \{S_{M_T} \neq \alpha_i\})\\
& \leq \mathbbm{P}(\lim_{M_T\to\infty} \{Z_{M_T} \neq P_{ij}\} ) +  \mathbbm{P}(\lim_{M_T\to\infty} \{S_{M_T} \neq \alpha_i\})\\
&= 0 + 0 = 0
\end{split}
\end{equation}

Since $\hat{P}_i^j = \frac{Z_{M_T}}{S_{M_T}}$ and $ P_i^j = \frac{p(y=i,\hat{y}=j)}{p(y=i)} = \frac{P_{ij}}{\alpha_i}$, we conclude $\hat{P}_i^j \xrightarrow{a.s.} P_i^j$. By using the same trick we can also prove $\hat{Q}_i^j \xrightarrow{a.s.} Q_i^j$

\section{Proof of Theorem~\ref{thm:thm1} (See page~\pageref{thm:thm1})}

Under the assumption that $p(x|y) = q(x|y)$ and $P_i^j=Q_i^j$ the Equation (\ref{eq:1}) can be modified as

\begin{equation} \label{eq:13}
\begin{split}
J_{\gamma}&=\sum_{j=1}^{L}(\sum_{i=1}^{L}\gamma_{i}P_i^{j}-q(\hat{y}=j))^2 = \sum_{j=1}^{L}(\sum_{i=1}^{L}\gamma_{i}P_i^{j}-\beta_{i}Q_i^{j})^2\\
&=\sum_{j=1}^{L}(\sum_{i=1}^{L}\gamma_{i}P_i^{j}-\beta_{i}P_i^{j})^2 = \sum_{j=1}^{L}(\sum_{i=1}^{L}(\gamma_{i}-\beta_{i})P_i^{j})^2
\end{split}
\end{equation}

Obviously, $J_{\gamma} \geq 0$ and the equality is satisfied when $\gamma = \beta$. Let $\bar{q} = [q(\hat{y}=1),q(\hat{y}=1),...,q(\hat{y}=L)]^T$, the Equation (\ref{eq:1}) can also be expressed as

\begin{equation}
J_{\gamma} = ||\bar{P}\cdot\gamma - \bar{q}||^2  \label{eq:14}
\end{equation}

The Hessian matrix of $J_{\gamma}$ is $2\bar{P}^T\bar{P}$, which is a positive semidefinite matrix. Thus we conclude that $J_{\gamma}$ is convex.

\end{document}